\documentclass[letterpaper, 10 pt, conference]{IEEEtran}

\IEEEoverridecommandlockouts                              

\usepackage{graphics} 
\usepackage{epsfig} 
\usepackage{mathptmx} 
\usepackage{times} 
\usepackage{amsmath} 
\usepackage{amssymb}  
\usepackage{algorithm}
\usepackage{algpseudocode}
\usepackage{amsmath}
\usepackage{tabularx}
\usepackage{booktabs}   
\usepackage{siunitx}    
\usepackage{threeparttable}
\usepackage{textcomp}

\usepackage{graphicx}
\let\labelindent\relax 
\usepackage{enumitem}
\usepackage{caption}
\usepackage{subcaption}
\usepackage{url}
\usepackage{xcolor}
\usepackage{amssymb}
\usepackage{pifont}
\usepackage{amssymb}
\usepackage{booktabs}
\usepackage{tabularx}
\usepackage{tikz} 
\usepackage{microtype} 
\usepackage{pifont}
\usepackage{xcolor}
\usepackage{tabularx}
\usepackage{amsmath}
\usepackage{adjustbox}

\usepackage{booktabs} 
\usepackage{siunitx} 
\usepackage{multirow} 
\usepackage[T1]{fontenc}

\title{\LARGE \bf
Graph-Based Adaptive Planning for Coordinated Dual-Arm Robotic Disassembly of Electronic Devices (eGRAP)
}

\author{Adip Ranjan Das$^{1}$ and Maria Koskinopoulou$^{2}$
\thanks{*This work was not supported by any organization}
\thanks{$^{1}$Adip Ranjan Das, School of Engineering and Physical Sciences, Heriot-Watt University, Edinburgh, UK
        {\tt\small ard2000@hw.ac.uk}}%
\thanks{$^{2}$Maria Koskinopoulou, School of Engineering and Physical Sciences, Heriot-Watt University, Edinburgh, UK
        {\tt\small M.Koskinopoulou@hw.ac.uk}}%
}

\begin{document}

\maketitle
\thispagestyle{empty}
\pagestyle{empty}

\begin{abstract}
E-waste is growing rapidly while recycling rates remain low. We propose an electronic-device Graph-based Adaptive Planning (eGRAP) that integrates vision, dynamic planning, and dual-arm execution for autonomous disassembly. A camera-equipped arm identifies parts and estimates their poses, and a directed graph encodes which parts must be removed first. A scheduler uses topological ordering of this graph to select valid next steps and assign them to two robot arms, allowing independent tasks to run in parallel. One arm carries a screwdriver (with an eye-in-hand depth camera) and the other holds or handles components. We demonstrate eGRAP on \SI{3.5}{in} hard drives: as parts are unscrewed and removed, the system updates its graph and plan online. Experiments show consistent full disassembly of each HDD, with high success rates and efficient cycle times, illustrating the method’s ability to adaptively coordinate dual-arm tasks in real time.
\end{abstract}

\IEEEkeywords 
Robotic Disassembly, 
Dual-Arm Manipulation, Task Planning,
Sustainable Manufacturing

\section{Introduction}

Electronic waste (e-waste) continues to grow while formal collection and recycling rates remain low. This leads to loss of valuable materials and increased environmental risk when hazardous substances enter the waste stream \cite{ewastemonitor_global_2024}\cite{poschmann_disassembly_2020}. Robotic disassembly offers a controlled alternative to shredding by taking devices apart in a defined order so parts and materials can be sorted and recovered.

Many existing robotic cells are built for a single product and follow a fixed program. These systems work when the device and environment are tightly controlled, but they struggle when parts vary, fasteners are missing, or components are damaged. Industrial efforts illustrate both the potential and the limits of specialisation: Apple’s iPhone recycling robots and automated hard drive processing initiatives achieve high throughput by exploiting product uniformity \cite{rujanavech_liam_nodate}\cite{noauthor_apple_nodate}\cite{microsoft,noauthor_microsoft_nodate}. However, such lines allow little adaptation and cannot easily exploit parallel work.

\begin{figure}[t]
    \centering
    \includegraphics[width=0.65\linewidth]{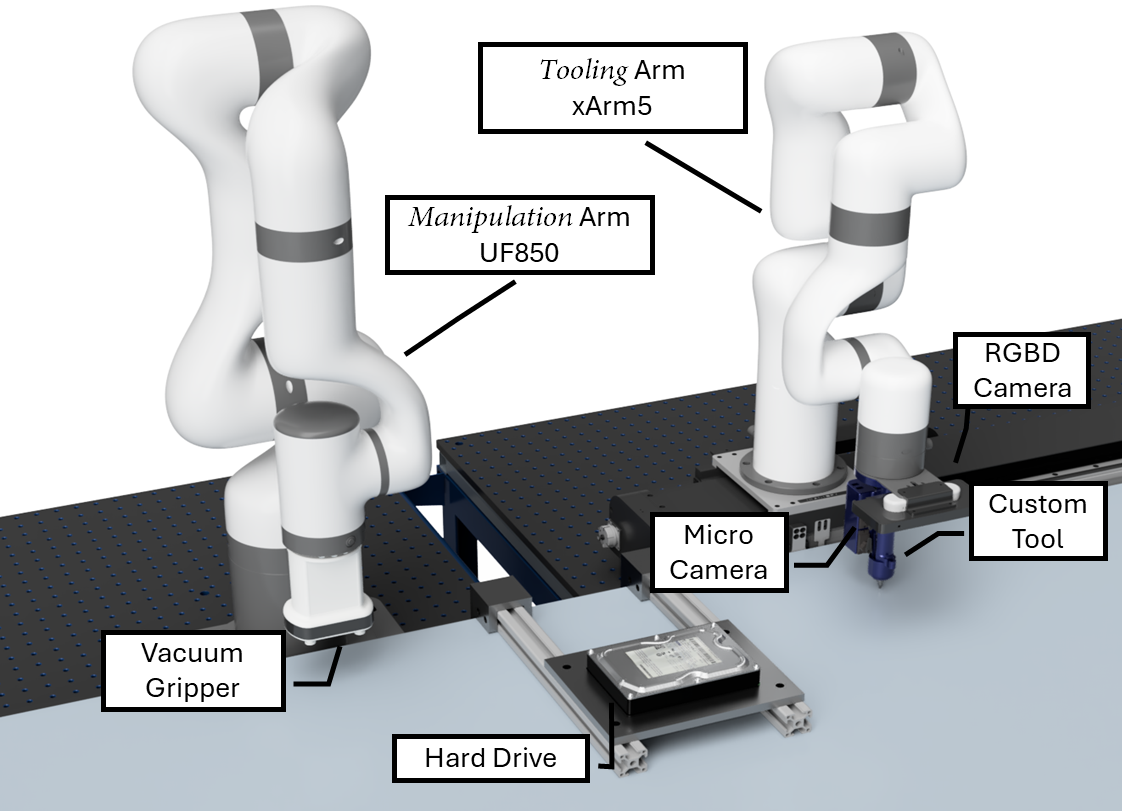}
    \caption{Dual-Arm testbed with a \emph{Manipulation} arm (vacuum gripper) and a \emph{Tooling} arm (screw-driving tool, RGB–D camera, and micro-camera). A \SI{3.5}{in} HDD is fixed on a passive holder at the workspace centre.}
    \label{fig:testbed}
\end{figure}

A flexible system needs perception, planning, and execution that can change with the scene. After detecting parts, it is natural to model the device as a directed graph of components with edges for precedence and access, and to compute a valid order by topological reasoning \cite{jungbluth_intelligent_2017}\cite{cui_robotic_2023}\cite{kiyokawa_many-objective-optimized_2024}. Updating this graph online when observations change (for example, a missing or newly revealed fastener) allows the sequence to adapt during execution. Recent reviews on robotic assembly and disassembly motivate such integrated and adaptive approaches for sustainable manufacturing \cite{ranjan_das_toward_2025}. Using two robots further increases capability: one arm can hold or stabilise while the other operates, or both can work on independent targets to reduce idle time \cite{buhl_dual-arm_2019}\cite{poschmann_disassembly_2020}. At the action level, screw removal is sensitive to small pose errors and surface reflections, so close-range visual alignment and contact checks are needed before applying torque \cite{szewczyk_uncertainty_2024}.

This paper proposes \emph{electronic-device Graph-based Adaptive Planning (eGRAP)} for autonomous disassembly. The method builds a directed part graph from live RGB-D detections, encodes precedence and access rules, and uses a topological scheduler to select the next valid actions. A coordination scheme assigns actions to two arms under collision and workspace limits and exposes safe parallel steps when dependencies allow it. The plan updates online after each step so the system can react to new observations or partial failures. The framework is device-agnostic: changing the object labels, detector training, and rules is sufficient to apply the same reasoning and scheduling to a new product. The proposed approach is tested in real-world experiments with two robotic arms on three different hard drive models (Figure \ref{fig:testbed}), completing full teardowns within an average of 22 mins per unit and demonstrating robust perception, adaptive planning, and reliable dual-arm execution.

The contributions address gaps in the state of the art. First, we present a unified formulation that links perception, graph construction, topological sequencing, and coordinated dual-arm execution in one loop. Second, we introduce online maintenance of the precedence graph from live detections and immediate rescheduling after each step, which provides adaptation to variation and missing parts. Third, we define a dual-arm scheduling and coordination scheme that respects access and collision constraints while enabling hold--operate behaviours and independent-step execution. Fourth, we include a fastener-engagement routine that combines close-range visual alignment with contact validation to address uncertainty in screw removal \cite{szewczyk_uncertainty_2024}. Compared to prior systems that are single-arm, device-specific, or rely on fixed scripts \cite{poschmann_disassembly_2020}, eGRAP targets autonomous disassembly of electronic devices with shared tasks across two arms and with adaptation driven by perception and graph reasoning. A full video demonstration of the complete Samsung HDD disassembly is available at: \url{https://www.youtube.com/watch?v=7a6v7ff-EvU}.


\section{Related Work}

Robotic disassembly has progressed along two complementary lines: concrete systems that dismantle real devices under sensing and control constraints, and process frameworks that formalize how to perceive, plan, and execute disassembly tasks under variability. A representative example of system-level integration is the vision-guided unscrewing cell by Díaz \textit{et al.}, which combines RGB-D detection of screw-head types with a mechatronic screwdriver and force-aware alignment to automate fastener removal on end-of-life products \cite{diaz_robotic_2025}. Kranz \textit{et al.} document lessons from the Robothon Grand Challenge, reporting a modular Sense-Plan-Act architecture, board localisation accuracy, and task execution time on real e-waste boards, with comparisons to human benchmarks \cite{kranz_robothon_2023}. In electric-vehicle battery recycling, Peng \textit{et al.} introduce BEAM-1, a mobile manipulator that employs neuro-symbolic AI and heuristic search to plan and execute bolt-removal sequences; extensive trials show sustained autonomous disassembly of multi-bolt modules \cite{peng_beam1_2024}. Complementary domain studies describe multi-robot battery-pack disassembly platforms with quantitative throughput and success metrics, demonstrating transfer to varied pack designs and fixtures \cite{qu_battery_platform_2024}.

Several works propose process and data frameworks for scalable automation. Saenz \textit{et al.} survey digital-twin and data-modelling requirements for automatic electronics disassembly, arguing for explicit representations of product states, skills, and step dependencies derived from both top-down analysis and bottom-up manual teardowns \cite{saenz_automated_2024}. Asif \textit{et al.} review task-and-motion planning (TAMP) for end-of-life products, emphasising that dynamic sequence generation and re-planning are required to cope with uncertain part condition and product variants, especially for EV batteries \cite{asif_tamp_2024}. These perspectives converge on the need to link perception outputs to a structured task representation that can be updated online.

Human-robot collaboration and tele-robotics provide additional evidence that coordination and explicit scheduling improve robustness. Ameur \textit{et al.} present a human-robot collaborative smartphone disassembly line in which operators can reassign tasks via voice/gesture control; simulated trials show reduced cycle time with low error rates when mixed-initiative task sharing is enabled \cite{ameur_hrc_2025}. For hazardous battery work, Hathaway \textit{et al.} compare asymmetric haptic master-slave control to symmetric dual-cobot teleoperation during module unbolting and cutting, finding 22-57\% time reductions for the symmetric setup at a modest cost to first-attempt success rates \cite{hathaway_frontiers_nodate}. These studies highlight how coordinated multi-agent action and guarded control can sustain throughput under uncertainty.

Perception for disassembly focuses on small, reflective, and densely packed targets. Yildiz \textit{et al.} demonstrate a hybrid 2D/3D pipeline for HDDs that localizes screws, lid, and pry points, noting the value of close-range imaging for metallic surfaces \cite{yildiz_visual_2020}. In parallel, edge/IoT pipelines with modern detectors have been explored for PCB-level component detection to support online decision making in disassembly lines \cite{mohsin_edgepcb_2025}. Across these efforts, reliable fastener engagement remains a critical step: uncertainty-aware routines combining visual alignment and contact checks help prevent slip and damage during unscrewing, which is now a frequent design feature of disassembly cells \cite{diaz_robotic_2025}.

Sequence generation is commonly formalised on graphs. Directed graphs with precedence and access edges enable topological ordering for feasible removal sequences, and they can be extended with objectives or uncertainty, as surveyed in recent TAMP and digital-twin literature \cite{asif_tamp_2024}\cite{saenz_automated_2024}. Competition and case-study reports also encode tasks as dependency graphs to expose parallelism and to support re-planning when detections change \cite{kranz_robothon_2023}\cite{qu_battery_platform_2024}. This growing body of work motivates frameworks that (i) convert live detections into a part graph, (ii) compute precedence-feasible orders that expose independent work, and (iii) coordinate execution—potentially with two arms—to reduce idle time while respecting access and safety constraints.

Prior studies provide strong components: close-range vision and contact-aware unscrewing \cite{diaz_robotic_2025}\cite{yildiz_visual_2020}, graph-based sequencing with re-planning hooks \cite{asif_tamp_2024}\cite{saenz_automated_2024}, and coordinated multi-agent execution for efficiency and safety \cite{hathaway_frontiers_nodate}\cite{ameur_hrc_2025}. What remains less explored is a single, device-agnostic loop that tightly couples live perception to an online-updated precedence graph and a dual-arm scheduler that can both synchronise hold-operate behaviours and run independent steps in parallel, a gap that the present work addresses with eGRAP.


\section{Electronic-Device Graph-based Adaptive Planning (eGRAP) Framework}
\label{sec:egrap}

\begin{figure}[t]
    \centering
    \includegraphics[width=0.8\linewidth]{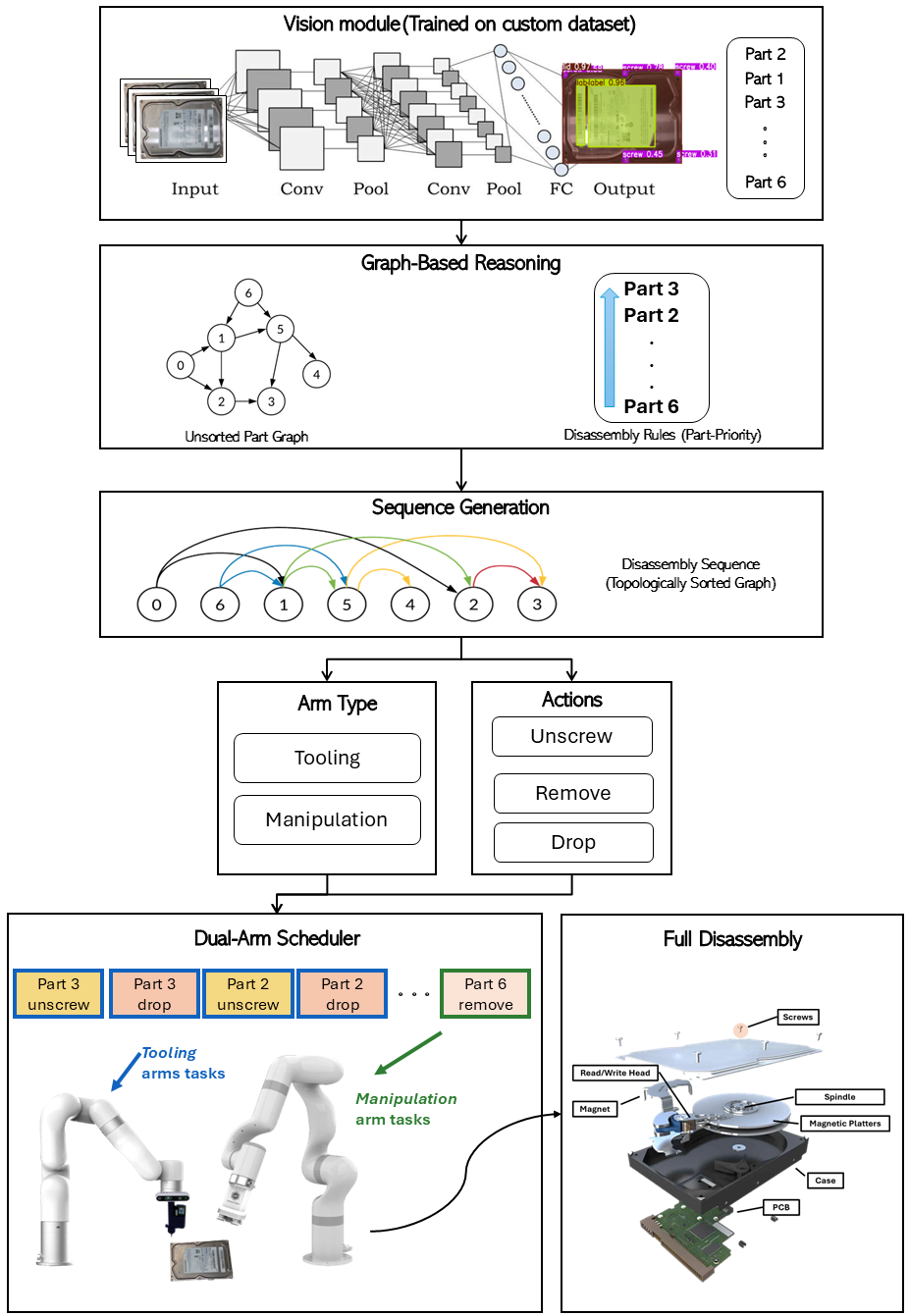}
    \caption{Block diagram of the eGRAP framework. \textit{Perception} outputs labeled parts and 3D poses. The \textit{Graph} block encodes precedence and access rules. The \textit{Sequence Generator} topologically orders ready parts and instantiates primitive actions. The \textit{Scheduler} assigns actions to two arms and updates the plan online from new observations. In the timeline, \textbf{blue} outlines denote the tooling (screwdriver) arm and \textbf{green} outlines denote the manipulation (gripper) arm; fill colors denote action types: \textbf{yellow} = \emph{unscrew}, \textbf{purple} = \emph{remove}, \textbf{orange} = \emph{drop}.}
    \label{fig:egrap_block}
\end{figure}

The eGRAP framework is a closed-loop system for robotic disassembly of electronic devices. It combines perception (to detect and identify parts), graph-based reasoning (to decide a removal order), and dual-arm execution (two robot arms coordinating to perform actions). The approach generalizes across devices and robot setups: to transfer to a new device family, one only updates the set of part types, a compact set of precedence/access rules, and a library of action templates; the core graph model, sequence generator, and scheduler remain unchanged.
The complete eGRAP framework, showing the vision module, graph-based reasoning, sequence generation, and dual-arm scheduler, is illustrated in Figure~\ref{fig:egrap_block} and described in the following.



\subsection{Vision Module}
The perception pipeline has two stages that work together. First, the eye-in-hand RGB-D camera provides a global view of the device. On each frame, the visible parts are detected and classified based on the learned classes (for HDDs: \emph{screw}/\emph{fastener}, \emph{lid}, \emph{PCB}), giving the confidence score, and the 2D pixel location of the detected part. Each 2D detection is projected into 3D by applying the calibrated camera intrinsics and sampling the corresponding depth value at the detection centroid \cite{calib}. The resulting 3D points are then transformed into the shared world coordinate frame using the hand–eye calibration, providing target poses that can be directly consumed by the planning and motion modules. This procedure is applied to all part types and supplies the coarse approach poses for the robots.

The second stage is a local, close-range refinement used only for screws. After the tooling arm reaches the coarse pose over a selected fastener, this fine manipulation process is employed. The same detection model runs on streams captured by a micro-camera to give a close view to detect screw-heads precisely. The pixel offset between the detected recess centre and the camera’s optical axis gives a small in-plane correction, which the arm applies before seating the driver. 
Because screw heads are small, reflective, and depth can be noisy on metal, this two-stage approach with a global RGB-D for coarse positioning and a micro-camera for fine alignment, allows fine alignment and centres the driver bit in the T8 recess before torque is applied. Larger parts such as lids and PCBs are well handled with the RGB-D view, so they use only the first stage. 

The dection model used in both cases is YOLOv11 \cite{yolo11_ultralytics} trained and fine-tuned on a custom dataset of 250 images that we manually annotated in Roboflow \cite{roboflow}. Augmentations including horizontal/vertical flips, \(\pm\ang{15}\) rotations, \(\pm\ang{10}\) shears, \(\pm 20\%\) brightness change, light blur \(\le\)1\,px, and sparse noise \(\le 0.97\%\) pixels), were added \emph{offline in Roboflow} and the augmented set was exported for training, yielding roughly 1{,}300 effective images. Images are auto-oriented and resized to \(640\times 640\) for training and inference.
Table~\ref{tab:train_cfg} summarises the configuration settings of the detection model.

Detectors are trained per device family to remain sensitive to small or shiny parts. Across frames, we associate detections by simple class-consistent nearest-neighbour matching in image space; this maintains a short track for each physical item. When two detections refer to the same item in a frame (for example, due to overlapping boxes), we merge them if they fall within a small spatial gate and average their positions. The resulting 2D centroids are lifted to 3D by back-projecting with the calibrated camera intrinsics and using the depth map at the centroid (with a small neighbourhood median for robustness). Hand-eye calibration provides the transform from the eye-in-hand camera to the robot/world frame, so 3D poses are expressed in the shared world frame used by planning and execution. A light exponential moving average smooths the 3D positions over time to reduce jitter.

The perception output is a set of labelled part instances with 3D poses. When a previously unseen item appears (e.g., internals revealed after lid removal) or a known item disappears (e.g., a screw is extracted), the system updates the planning graph immediately. This keeps the plan consistent with the scene and allows the sequence to adapt as the disassembly progresses.

\begin{table}[h]
\centering
\caption{YOLOv11 training data and augmentation settings}
\label{tab:train_cfg}
\begingroup
\scriptsize
\setlength{\tabcolsep}{4pt}
\renewcommand{\arraystretch}{1.08}
\begin{tabularx}{\columnwidth}{@{}lX@{}}
\toprule
Base images & 250 (manually labelled in Roboflow) \\
Effective images (offline aug., exported) & $\approx$1{,}300 \\
Auto-orient & Applied \\
Resize & Stretch to $640\times 640$ \\
Augmentations (offline, Roboflow) & H/V flip; rotation $\pm 15^{\circ}$; shear $\pm 10^{\circ}$ (H/V); brightness $\pm 20\%$; blur $\le 1$\,px; noise $\le 0.97\%$ of pixels \\
\bottomrule
\end{tabularx}
\endgroup
\vspace{-2mm}
\end{table}

\subsection{Graph-Based Reasoning and Sequencing}
The \emph{Part Graph} represents the current device state using common graph terms. Each detected item is a \emph{node}. A directed \emph{edge} between two nodes encodes a rule that one part must be cleared before the other (precedence) or that one part blocks access to the other (access). A compact set of class-level rules automatically adds the correct edges; for example, “all \emph{fastener} nodes point to their \emph{lid} node,” and “the \emph{lid} points to any internal components”. When a node is confirmed removed, it is dropped from the graph along with its incident edges, so the graph always reflects the visible, current device.

Given the graph, the sequence generator builds a \emph{precedence-feasible order} of parts to remove. It first collects the \emph{ready set} of nodes: these are parts that are still present and have no incoming edges, meaning nothing else is blocking them. If several nodes are ready at the same time, simple tie-breakers are applied: class priority (fasteners before lid, then internal components, then case bottom side parts) and a short-move preference so nearby targets are grouped. The ordered nodes are then turned into \emph{action primitives} that the robots can execute, such as \emph{unscrew}, \emph{lift}, \emph{remove}, and \emph{drop}. Each action is created from a template with approach offsets, grasp/seat poses, nominal speeds, and tool parameters. Preconditions are checked against the graph—for example, a \emph{lift} action for a lid is only issued when all its incident fasteners no longer appear in the graph.

\subsection{Dual-Arm Calibration and Scheduling}
Before execution, both arms are brought into the same world frame using a short, vision-based calibration. An ArUco marker is placed on the work surface. The eye-in-hand camera on the tooling arm is used to drive the tool centre point (TCP) to the marker centre, and small corrective motions are applied until the measured pose aligns with the known marker pose. The manipulation arm is then jogged to the same physical point and fine-tuned, so both arms agree on a common coordinate frame within a small tolerance.

Each action declares the \emph{capability} it needs: \emph{unscrew} requires the tooling arm (screwdriver), while \emph{lift}, \emph{remove}, and \emph{drop} require the manipulation arm (gripper). The scheduler looks at the current ready actions, filters them by capability, and then selects a set that can run \emph{in parallel} without spatial interference. Two actions are considered non-interfering when their workspaces do not overlap and they have no direct dependency in the graph. Non-interfering actions are dispatched concurrently to the two arms; overlapping or dependent actions are synchronised to realise safe \emph{hold-operate} behaviours and to minimise idle time.

Fastener handling includes a brief engagement check inside the \emph{unscrew} primitive. The driver is visually aligned over the screw head, a short guarded axial motion confirms seating, and only then is rotation started. If engagement is not confirmed, the action is re-queued with a small pose adjustment while other ready actions continue. When an action finishes successfully, its node is removed from the graph and newly unblocked nodes become ready immediately.

\subsection{Execution Loop}
eGRAP runs in a fixed-rate loop: perception updates detections; the graph is refreshed from class rules; a topological pass exposes the next ready actions; actions are instantiated from templates and scheduled; execution feeds back state. As parts are removed or revealed, the ready set and schedule update immediately. Because the planner operates on abstract labels and capabilities, the same loop transfers across products and robot platforms: swapping detectors, adding device rules, or refining templates does not change the core pipeline of converting live detections to a precedence graph, generating a valid order, and executing it with coordinated dual-arm control.


\section{Experimental Results}
\label{sec:results}

\subsection{Testbed and System Configuration}
\label{subsec:testbed}

The experimental setup consists of two collaborative robot arms operating within a shared workspace.

\begin{figure}[b]
    \centering
    \includegraphics[width=0.69\linewidth]{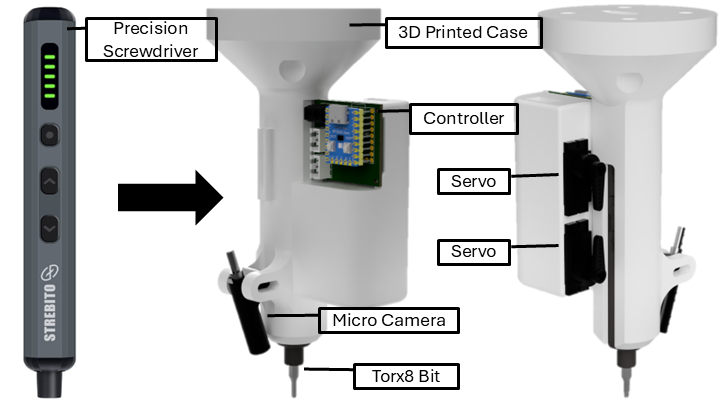}
    \caption{Custom end-effector with a \emph{Soleilwear} precision electric screwdriver, a 2D RGB micro-camera, two micro-servos for forward/reverse, and a controller. A Torx~T8 bit is used.}
    \label{fig:tool}
\end{figure}

The \textbf{\emph{Tooling} arm}, a 5-DoF xArm5, is equipped with a custom-made screw-driving end-effector and an Intel RealSense D435i in an eye-in-hand configuration. The RGB-D stream from this sensor is processed by eGRAP for part detection and close-range alignment prior to screw engagement. A micro-camera is embedded inside the tool body and aimed along the driver’s axis toward the screwdriver \emph{bit}—that is, the removable Torx~T8 screwdriver bit that engages the screw head. The camera has a resolution of $640{\times}480$ (0.3\,MP) at 30\,fps, a $\sim\!67^{\circ}$ field of view, and a close-focus range of 3-10\,cm. This placement gives a clear, close-up view of the screw and the Torx~T8 bit during approach. The image is used to make small in-plane and axial corrections so the bit seats fully in the screw head, and to visually confirm contact before starting rotation for the unscrewing step.
The screw-driving tool is built on a commercial precision electric screwdriver by \emph{Soleilwear}, mounted inside a 3D-printed housing case (Figure~\ref{fig:tool}). The housing provides the wrist-flange interface and keeps all auxiliary components rigid so that the bit axis is aligned to the tool frame. To avoid modifying the commercial driver electronics, two miniature servos press the driver’s physical buttons: one for the \emph{forward} (screw) button and one for the \emph{reverse} (unscrew) button. Short servo pulses emulate a human press, allowing the robot to switch modes and start/stop rotation without any internal changes to the driver.

The \textbf{\emph{Manipulation} arm}, a 6-DoF UF850 with an electric vacuum gripper, stabilises the device during tooling, removes freed components such as the lid or PCB, and deposits them into collection bins. Coordination between the two arms is managed by the eGRAP scheduler, which enables parallel execution of independent tasks while ensuring safe hold-operate behaviours. 
The complete testbed, including both robotic arms, end-effectors, sensors, and supporting hardware, is shown in Figure~\ref{fig:testbed}. The device under test is a \SI{3.5}{in} hard drive. 

\begin{figure}[t]
    \centering
    \begin{subfigure}{0.6\linewidth}
        \centering
        \includegraphics[width=\linewidth]{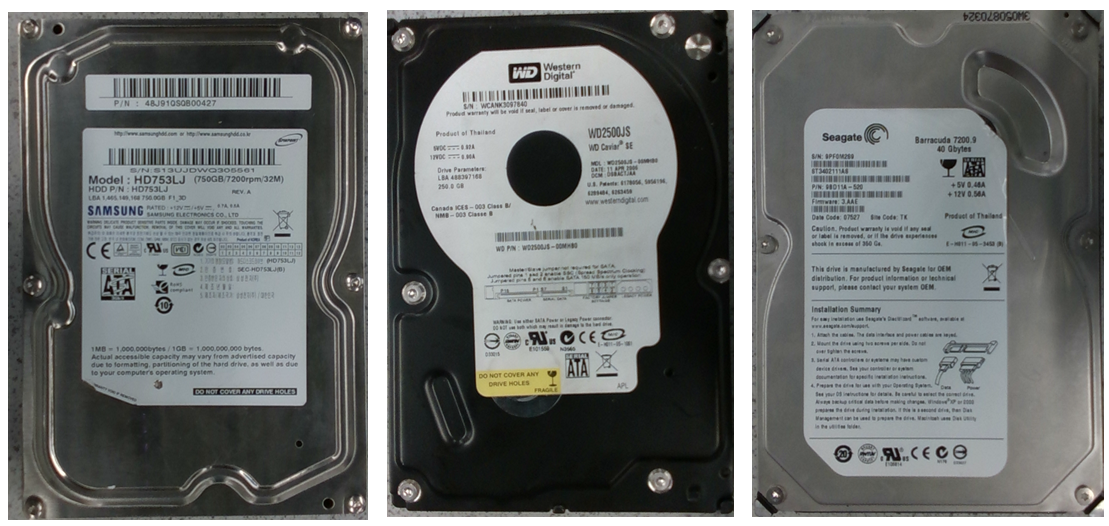}
        \caption{}
    \end{subfigure}
    \vspace{1mm}
    \begin{subfigure}{0.19\linewidth}
        \centering
        \includegraphics[width=\linewidth]{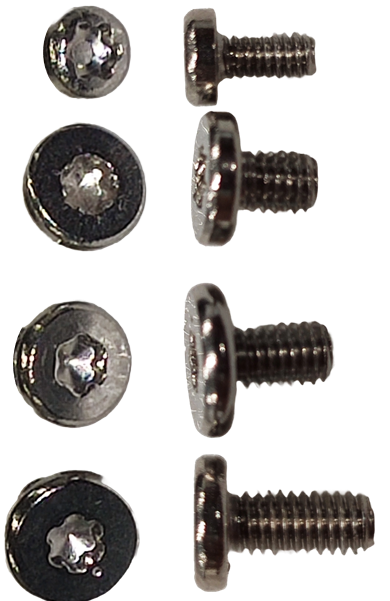}
        \caption{}
    \end{subfigure}
    \caption{HDD brands and fasteners.}
    \label{fig:hdd_family_screws}
\end{figure}


The hard drives used in our experiments are \SI{3.5}{in} units from three major brands, i.e. Samsung, Western Digital, and Seagate, as shown in Figure~\ref{fig:hdd_family_screws}a. Across these brands, the drives contain four variants of Torx~T8 screws (Figure~\ref{fig:hdd_family_screws}b). These variants differ in head diameter and shank length, but all share the same T8 drive geometry. In a full teardown we encounter seven T8 screws on the lid (L1), twelve internal T8 screws (L2), and five T8 screws on the case bottom side (L3). Accordingly, the tool uses a single Torx~T8 bit for all brands and all layers.
For all experiments, the HDD is mounted on a passive holder at the centre of the workbench. This fixture constrains the drive, exposes the top surface for access to lid screws, and provides sufficient clearance for the manipulation arm to lift components once released by the tooling arm. 

Each experiment runs the full eGRAP loop with the two robots coordinating to share tasks. The \emph{Tooling arm} moves to a coarse pose from RGB-D perception, performs fine alignment on a Torx~T8 screw using the embedded micro-camera, and actuates the screwdriver to unscrew, retrying if engagement is lost. Freed screws are carried to a drop location. In parallel, the \emph{Manipulation arm} holds or lifts components once all constraining fasteners are cleared and places them in designated bins. The planner maintains a live part graph, updating nodes and the ready set after each action, while the scheduler assigns non-conflicting tasks to the two arms. Disassembly proceeds across three layers: L1 (lid and seven screws), L2 (twelve internal screws, platter holder, platter, and actuator), and L3 (five case bottom side screws, PCB, and case). Completion is defined as all nodes cleared and all parts for each layer successfully disassembled. To evaluate accuracy and repeatability, we performed 10 full teardowns per HDD family, calculating metrics as reported in the following, including detection accuracy, screw-level success, disassembly layer times, total cycle time, and full-teardown completion rate.

\subsection{Vision System Performance}
\label{subsec:vision-metrics}
The vision system was evaluated online during disassembly across the 10 iterations of the experiment for each three HDD families. Performance was measured using standard detection metrics, including precision (the proportion of correct detections), recall (the proportion of ground-truth instances detected), mean average precision at 0.5 IoU (mAP@0.5), and mean 2D localisation error in pixels on $640\times 640$ frames. Each metric was averaged over ten complete trials for each HDD family. The results are summarised in Table~\ref{tab:yolo_metrics}.

\begin{table}[h!]
\centering \caption{YOLOv11 detection metrics}
\label{tab:yolo_metrics}
\begingroup
\scriptsize
\setlength{\tabcolsep}{3pt}
\renewcommand{\arraystretch}{1.05}
\begin{tabular}{lcccc}
\textbf{Model} & \textbf{Precision} & \textbf{Recall} & \textbf{mAP@0.5} & \textbf{Loc.\ Err.\ (px)} \\
\midrule
Samsung         & 0.92 & 0.88 & 0.91 & 6.3 \\
Seagate         & 0.89 & 0.86 & 0.88 & 7.1 \\
Western Digital & 0.93 & 0.89 & 0.92 & 5.8 \\
\bottomrule
\end{tabular}
\endgroup
\vspace{-2mm}
\end{table}

The vision detection module maintained consistently high performance across all three HDD families. Western Digital exhibited the most accurate localisation (5.8 px) and the highest mAP@0.5 (0.92), closely followed by Samsung (mAP@0.5: 0.91, localisation error: 6.3 px). Seagate showed slightly lower recall (0.86) and mAP@0.5 (0.88), with a modest increase in localisation error (7.1 px). In practice, these differences affect how many fine-alignment iterations are needed before the screwdriver seats, but they do not change the computed precedence graph or the chosen sequence.

High precision limits false positives, which stabilises the part graph by avoiding spurious nodes. Recall near 0.9 means most fasteners and lid are detected early; any missed items are usually observed on the next view and inserted without violating precedence rules. The mean localisation error of about 6 px at $640\times 640$ is within the capture range of the fine manipulation routine, which explains the observed behaviour in the lid layer: coarse positioning alone is fast but misses engagement on several screws; with fine manipulation, all screws are cleared at the cost of additional time (see Table~\ref{tab:l1_coarse_fine}). 

Rare perception faults are linked to surface reflections on some lids. When this occurs, the plan pauses on the affected node, a short re-scan is triggered, and the graph is updated with the new detections. Because eGRAP separates coarse detection from contact-validated engagement, these transient errors have limited impact on overall success and do not require changes to the sequence.

\subsection{HDD Sequence Generation with eGRAP}
\label{subsec:sequence-gen}

\begin{figure}[t]
    \centering
    \includegraphics[width=0.7\linewidth]{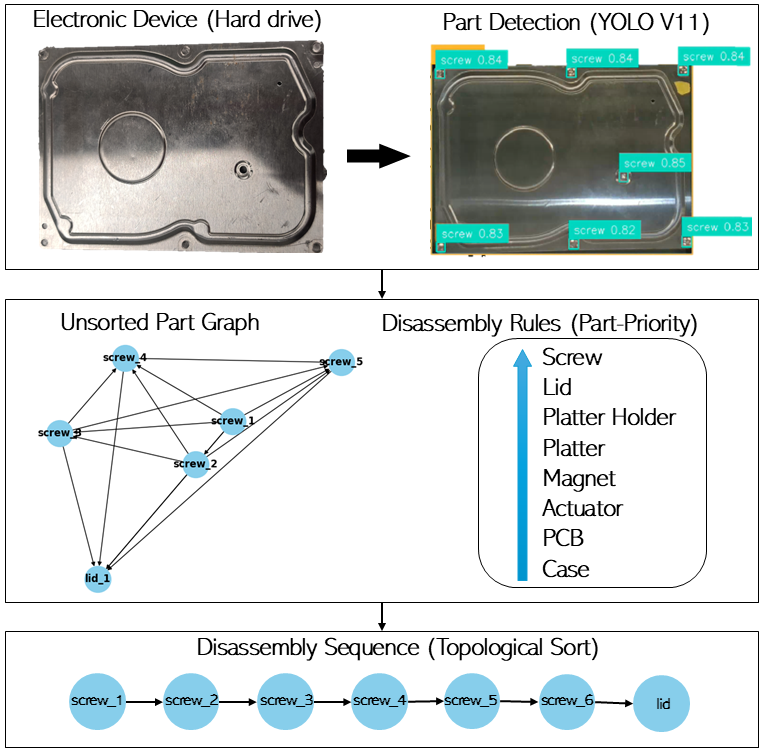}
    \caption{HDD sequence generation with eGRAP (L1 example). Top: the detector finds \emph{screw} and \emph{lid}. Middle: detections become nodes; class rules add edges (e.g., \emph{screw} $\rightarrow$ \emph{lid}). Bottom: a topological sort yields a valid order for lid removal. L2 and L3 follow the same pattern with different labels and rules.}
    \label{fig:seqgen}
\end{figure}

We apply eGRAP to \SI{3.5}{in} hard drives from the three families using one shared list of part labels and a small set of rules. The labels are simply the part names the detector must find (e.g., screw, lid, PCB), and those same labels become the nodes that the planner reasons about in the graph. The same configuration is used across all three families; only geometry and layout differ. The framework builds and solves the sequence online as the scene changes so that small brand-specific differences are handled naturally.

Figure~\ref{fig:seqgen} illustrates the process for the lid layer (L1). The detector runs on the eye-in-hand RGB-D stream and returns labelled boxes for \emph{screw} and \emph{lid}. Detections are associated across frames, duplicates are merged by a small spatial gate, and 3D poses are estimated. Each confirmed instance becomes a node in the graph with class and pose.

\textit{L1 (lid).} For the lid layer, eGRAP adds edges from each \emph{screw} to the \emph{lid} (\emph{screw} $\rightarrow$ \emph{lid}) so that the lid is removable only after all incident screws are cleared. A topological sort computes a valid order. When several screws are ready, ties are broken by a fixed class priority (fasteners first) and a short move estimate so that nearby screws are serviced together. The ordered nodes are mapped to actions: each \emph{screw} becomes an \emph{unscrew} action on the tooling arm; the \emph{lid} becomes a guarded \emph{lift} on the manipulation arm once the graph confirms that all predecessor screws are gone. As actions succeed, their nodes are removed from the graph and successors become ready. Figure~\ref{fig:seqgen} shows this end-to-end flow for L1.

\textit{L2 (internals).} After lid removal, new parts become visible. The perception module adds nodes for \emph{platter\_holder\_screw}, \emph{actuator\_screw}, \emph{platter\_holder}, \emph{platter}, and \emph{actuator}. Class rules add edges from internal screws to their hosts (\emph{platter\_holder\_screw} $\rightarrow$ \emph{platter\_holder}, \emph{actuator\_screw} $\rightarrow$ \emph{actuator}) and edges from the \emph{lid} to all internal parts to model access already satisfied by L1. The same topological procedure selects the next valid steps. Internal screws are cleared first, then \emph{platter\_holder} and \emph{platter} are lifted in order, with the \emph{actuator} removed once its fasteners are cleared. As before, \emph{unscrew} actions are assigned to the tooling arm, and \emph{lift}/\emph{remove} actions to the manipulation arm. Independent targets that do not overlap in the workspace are run in parallel.

\textit{L3 (case bottom side).} The drive is then turned to expose the case bottom side. The detector adds nodes for \emph{case\_bottom\_screw}, \emph{PCB}, and \emph{case}. Rules enforce \emph{case\_bottom\_screw} $\rightarrow$ \emph{PCB} and \emph{PCB} $\rightarrow$ \emph{case}. The scheduler again favours fasteners first and short moves. After all \emph{case\_bottom\_screw} nodes are cleared, the \emph{PCB} is lifted, followed by the \emph{case}. As each action completes, the graph is updated and the ready set is recomputed, so the plan remains consistent with the current observations.

Across L1--L3, the logic is the same for all three HDD families: detections form nodes, class rules add edges, a topological sort yields a precedence-feasible order, and actions are issued to the two arms with capability and workspace checks. If an \emph{unscrew} action reports a failed engagement, it is re-queued with a small pose offset while other ready actions proceed, and the graph continues to unlock new steps. This common process allows a single configuration of eGRAP to handle Samsung, Western Digital, and Seagate units without brand-specific scripts.

\subsection{HDD Disassembly Execution}
\label{subsec:sequence-exec-cover}

\begin{figure}[!t]
    \centering
    \includegraphics[width=0.54\linewidth]{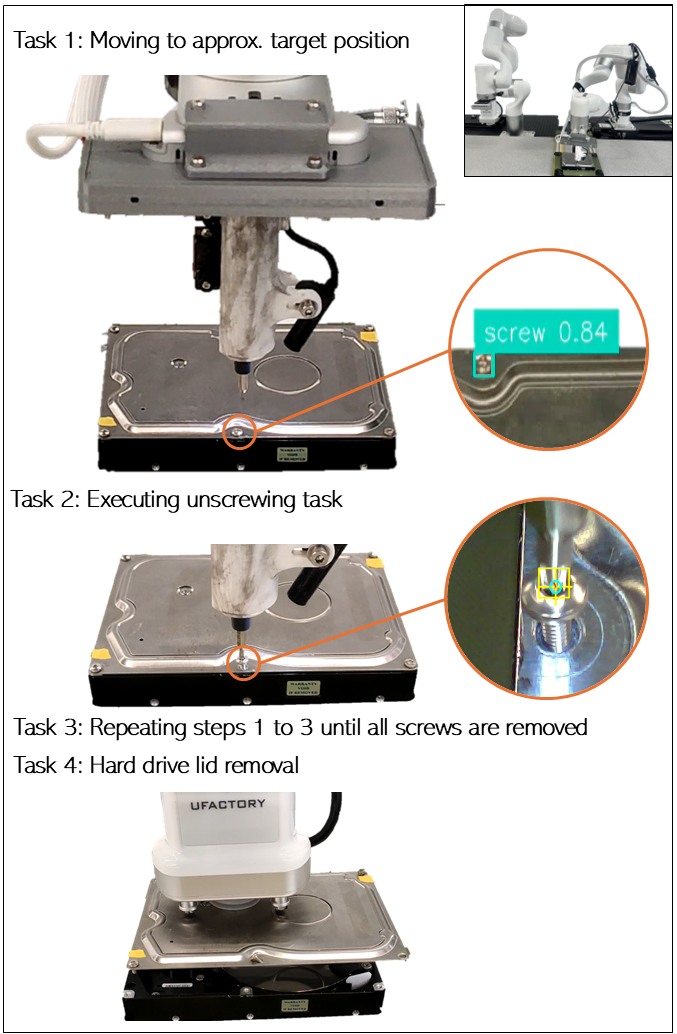}
    \caption{Execution for the lid layer (L1) with Indicative external view snapshot from the process (Top Corner). (1) Coarse move to the screw pose from RGB-D detection; (2) Unscrew with contact-validated engagement; (3) Repeat until all screws are cleared; (4) Lift and remove the lid with the manipulation arm.}
    \label{fig:seqexec_cover}
\end{figure}

\textit{Lid layer (L1).} Figure~\ref{fig:seqexec_cover} summarises the execution for the first layer. For every \emph{screw} node that becomes ready in the eGRAP plan, the scheduler issues an \emph{unscrew} action to the tooling arm and keeps a guarded \emph{lid lift} queued for the manipulation arm.

\emph{Coarse approach.} The tooling arm moves to an approach pose computed from the RGB-D detection and the calibrated eye-in-hand transform. The bit axis is aligned normal to the lid, and the arm stops a few millimetres above the target. This brings the tool into the neighbourhood of the screw head.

\emph{Engage and unscrew.} Small in-plane and axial corrections are applied to centre and seat the screwdriver bit in the Torx head. A short guarded motion verifies contact; if contact is not confirmed or a slip is detected, the arm backs off, applies a small offset, and retries. Once engagement is confirmed, a servo presses the driver’s button to rotate. A light axial preload is maintained during release. When time or depth change indicates that the thread is free, rotation stops, and the screw, captured at the bit, is carried to a drop pose and released into a bin.

The cycle repeats until all seven lid screws are cleared. When the graph confirms that no incident screws remain on the \emph{lid} node, the scheduler dispatches the queued \emph{lid lift}. The manipulation arm places the vacuum cup on a flat region, establishes suction, lifts the lid, and places it in the bin. L1 is completed, and the perception module exposes the internal parts, which are inserted into the graph for the next layer.

\begin{figure}[!t]
    \centering
    \includegraphics[width=0.54\linewidth]{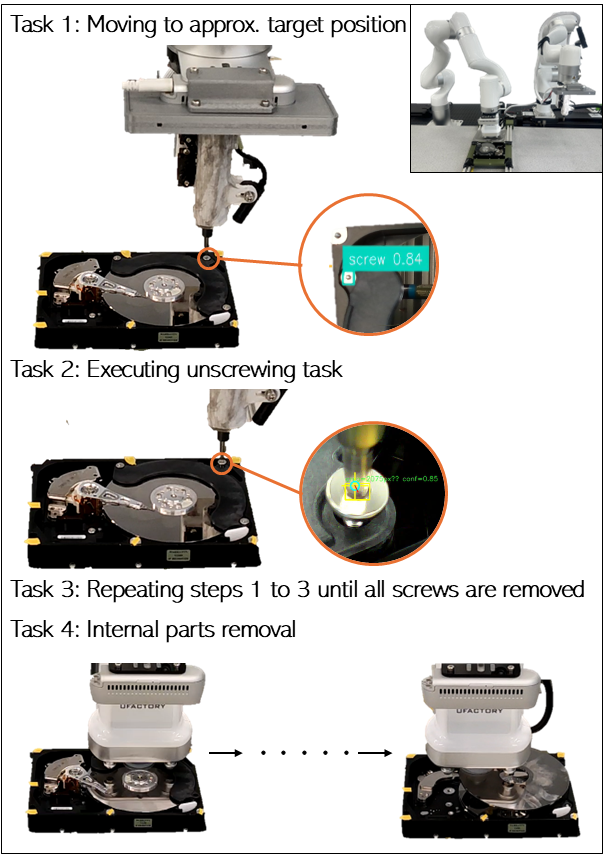}
    \caption{Execution for the internal layer (L2) with Indicative external view snapshot from the process (Top Corner). (1) Coarse move to the detected internal screw pose; (2) Unscrew with contact-validated engagement; (3) Repeat until all internal screws are cleared; (4) Remove internal components in order (platter holder, platter, actuator) using the manipulation arm.}
    \label{fig:seqexec_internal}
\end{figure}

\textit{Internal layer (L2).} After the lid is removed, the detector reveals twelve internal screws and components. Figure~\ref{fig:seqexec_internal} shows the same pattern applied to internal fasteners: coarse approach to each detected screw, contact-validated engagement and unscrewing, repetition until all internal screws are cleared, and then ordered removal of internal components. eGRAP enforces the graph rules so that the platter holder is lifted after its screws, followed by the platter and then the actuator. When targets are independent and their workspaces do not overlap, the scheduler runs actions in parallel to reduce idle time.

\textit{Case bottom side (L3).} The final layer follows the same procedure. Case bottom five screws are cleared first using the engage-unscrew cycle, then the PCB is lifted, and then the case is removed. The same eGRAP loop continues until the graph is empty.

\subsection{Disassembly Performance}
\label{subsec:disassembly-performance}

We evaluated eGRAP on three hard drive model families, performing ten full teardowns per family. The plan is organised into three layers: L1 (lid) removes seven screws and the lid; L2 (internals) removes twelve screws, the platter holder, the platter, and the actuator; L3 (case bottom side) removes five screws, the PCB, and the case. The scheduler issues parallel steps when targets are independent and their workspaces do not overlap. Table~\ref{tab:layer_times} reports mean layer times with fine manipulation enabled on L1; all totals are within $22$\,mins.

\begin{table}[t]
\centering
\caption{Layer-wise time for fine manipulation (mins)}
\label{tab:layer_times}
\begingroup
\scriptsize
\setlength{\tabcolsep}{3pt}
\renewcommand{\arraystretch}{1.05}
\begin{tabular}{lcccc}
\textbf{Model} & \textbf{L1} & \textbf{L2} & \textbf{L3} & \textbf{Total} \\
\midrule
Samsung         & 7.2 & 9.9 & 4.8 & 21.9 \\
Seagate         & 6.9 & 9.7 & 4.9 & 21.5 \\
Western Digital & 6.8 & 9.2 & 4.9 & 20.9 \\
\bottomrule
\end{tabular}
\endgroup
\vspace{-1mm}
\end{table}

L1 and L2 dominate the cycle since they are fastener-heavy. Western Digital is slightly faster overall (20.9\,mins), followed by Seagate (21.5\,mins) and Samsung (21.9\,mins). These small differences are consistent with minor variations in access and in the number of fine-alignment iterations needed before seating the bit.

To study the L1 speed-completeness trade-off, we compare coarse positioning only with fine manipulation using the micro-camera. Without fine manipulation, the tooling arm removes roughly three of seven screws quickly; with fine manipulation, all seven are cleared at the cost of extra time. Table~\ref{tab:l1_coarse_fine} reports the \emph{screw clearance rate (\%)} and the time for each model family.

\begin{table}[t]
\centering
\caption{L1: coarse only vs.\ fine manipulation (mean)}
\label{tab:l1_coarse_fine}
\begingroup
\scriptsize
\setlength{\tabcolsep}{3.2pt}
\renewcommand{\arraystretch}{1.05}
\begin{tabular}{lcccc}
\textbf{Model} & \multicolumn{2}{c}{\textbf{Coarse only}} & \multicolumn{2}{c}{\textbf{Fine manipulation}} \\
\cmidrule(lr){2-3}\cmidrule(lr){4-5}
 & \textbf{Clearance (\%)} & \textbf{Time (mins)} & \textbf{Clearance (\%)} & \textbf{Time (mins)} \\
\midrule
Samsung         & 42.9 & 2.1 & 100.0 & 7.2 \\
Seagate         & 42.9 & 2.0 & 100.0 & 6.9 \\
Western Digital & 42.9 & 2.0 & 100.0 & 6.8 \\
\bottomrule
\end{tabular}
\endgroup
\vspace{-1mm}
\end{table}

Table~\ref{tab:l1_coarse_fine} shows that fine manipulation guarantees full L1 clearance, preventing later delays. The extra time on L1 is recovered in L2 and L3 because the plan proceeds without backtracking and the scheduler keeps both arms busy.
Completion statistics for each model family are given in Table~\ref{tab:completion_counts}. We report the number of trials, the number of full teardowns completed, and the resulting success rate.

\begin{table}[t]
\centering
\caption{Full teardown completion}
\label{tab:completion_counts}
\setlength{\tabcolsep}{5pt}
\renewcommand{\arraystretch}{1.05}
\begin{tabular}{lccc}
\textbf{Model} & \textbf{Trials} & \textbf{Successes} & \textbf{Success rate (\%)} \\
\midrule
Samsung         & 10 & 9  & 90.0 \\
Western Digital & 10 & 7  & 70.0 \\
Seagate         & 10 & 9  & 90.0 \\
\midrule
Overall         & 30 & 25 & 83.3 \\
\bottomrule
\end{tabular}
\end{table}

Primary interruption causes were hardware or mechanical rather than planning-related. We observed safe aborts due to a transient vacuum-seal leak during a platter pick that led to part slip, intermittent reduction of tool-head illumination at the micro-camera that prevented reliable fine alignment during screw engagement, and instances where a lid screw was not fully disengaged before the guarded lid lift, triggering a resistance stop. In each case, the eGRAP plan remained valid; after correcting the condition or re-engaging the fastener, the same sequence executed to completion on re-run.
Taken together, Tables~\ref{tab:layer_times}-\ref{tab:completion_counts} show consistent full teardowns within $22$\,mins across all model families. The coarse-versus-fine comparison explains the L1 timing profile and aligns with the vision results: localisation errors of a few pixels are small enough for coarse approach but still benefit from fine alignment to guarantee first-pass screw removal. The scheduler’s parallelism reduces idle time by staging holds and lifts while the tool engages the next fastener, keeping totals tight across families and supporting stable end-to-end execution.


\section{Conclusion}
\label{sec:conclusion}

This paper presents eGRAP, a perception-driven, graph-based planning framework for coordinated dual-arm robotic disassembly of electronic devices. The method represents detected parts and their constraints as a directed graph, generates precedence-feasible sequences by topological sorting, and assigns actions to two arms under simple capability and non-overlap constraints. The framework closes the loop from perception to execution: new observations update the graph, the ready set is recomputed, and the scheduler issues actions accordingly. This design separates device content (labels, rules, and action templates) from the core reasoning and thus supports reuse across products and robot platforms.

We validated the approach on \SI{3.5}{in} hard drives from three manufacturers. The system completed full teardowns within 22\,mins per unit on average. The vision module, trained with modest data and targeted augmentations, maintained high precision and recall across families. The fine alignment routine, converted coarse visual poses into reliable screw engagement. The scheduler reads the precedence graph and each arm’s workspace to choose the next safe action. It assigns unscrew steps to the tooling arm and issues the matching hold or lift step to the manipulation arm only when the required conditions are met. Actions are sequenced so the arms never interfere and the device stays supported throughout.


The main contribution is a scalable disassembly pipeline that converts live detections into a precedence graph and applies simple rules to coordinate dual-arm actions. It generalises to new devices by updating part labels, retraining the detector, and redefining rules, while its instance-based reasoning allows it to handle missing or newly revealed parts without device-specific scripts.

Future work will expand the range of parts and products, including flexible elements like cables and gaskets, which need special perception and handling. A library of common part classes and subassemblies could be reused across devices, and using uncertainty estimates in the scheduler could help decide when to retry or reorder actions. eGRAP could also be integrated as a controller in larger recycling workflows, with support for job scheduling, traceability, and safety features to allow continuous operation. Finally, testing on standardized benchmarks with shared metrics would help compare systems and improve general robotic disassembly.



\bibliographystyle{ieeetr} 
\bibliography{references}

\end{document}